\documentclass{article}

     \usepackage[preprint]{neurips_2020}

\usepackage[utf8]{inputenc} 
\usepackage[T1]{fontenc}    
\usepackage[]{hyperref}       
\usepackage{url}            
\usepackage{soul}            
\usepackage{booktabs,tabularx,longtable}       
\usepackage[table]{xcolor}
\newcolumntype{Y}{>{\centering\arraybackslash}X}
\usepackage{amsfonts}       
\usepackage{nicefrac}       
\usepackage{microtype}      
\usepackage{graphicx}
\usepackage{multicol}
\usepackage{mathptmx}
\usepackage{amsmath}
\usepackage{amssymb}
\usepackage[export]{adjustbox}

\usepackage[ruled,vlined]{algorithm2e}

\title{Natural-Language Multi-Agent Simulations of Argumentative Opinion Dynamics}

\author{%
  Gregor Betz\\
  Karlsruhe Institute of Technology\\
  Karlsruhe, Germany\\
  \texttt{gregor.betz@kit.edu} \\
}

\begin{document}

\maketitle


\begin{abstract}
This paper develops a natural-language agent-based model of argumentation ({\smaller ABMA}). Its artificial deliberative agents ({\smaller ADA}s) are constructed with the help of so-called neural language models recently developed in AI and computational linguistics. {\smaller ADA}s are equipped with a minimalist belief system and may generate and submit novel contributions to a conversation. The natural-language {\smaller ABMA} allows us to simulate collective deliberation in English, i.e. with arguments, reasons, and claims themselves---rather than with their mathematical representations (as in formal models). This paper uses the natural-language {\smaller ABMA} to test the robustness of formal reason-balancing models of argumentation \citep{MaesFlache2013,Singer:2019oq}: First of all, as long as {\smaller ADA}s remain passive, confirmation bias and homophily updating trigger polarization, which is consistent with results from formal models. However, once {\smaller ADA}s start to actively generate new contributions, the evolution of a conservation is dominated by properties of the agents \emph{as authors}. This suggests that the creation of new arguments, reasons, and claims critically affects a conversation and is of pivotal importance for understanding the dynamics of collective deliberation. The paper closes by pointing out further fruitful applications of the model and challenges for future research.
\end{abstract}

\parskip=0em
\tableofcontents
\parskip=0.7em

\section{Introduction}

During the last decade, a variety of computational models of argumentative opinion dynamics have been developed and studied \citep[e.g.][]{Betz:2011DD,MaesFlache2013,olsson2013bayesian,Borg2017-BORAAA-7,Singer:2019oq,banisch2021}. These agent-based models of argumentation ({\smaller ABMA}s) have been put to different scientific purposes: to study polarization, consensus formation, or the veritistic value of argumentation; to understand the effects of different argumentation strategies, ascriptions of trustworthiness, or social networks; to provide empirically adequate, or epistemically ideal descriptions of joint deliberation. Moreover, {\smaller ABMA}s differ radically in terms of \emph{how} they represent argumentation, ranging from complex dialectical networks of internally structured arguments, to abstract argumentation graphs, to flat pro/con lists, to deflationary accounts that equate arguments with evidence. However, all these models are formal in the sense that they are built with and process mathematical representations of natural language arguments, reasons, claims, etc.---rather than these natural language entities themselves. 

\begin{figure}
  \centering
  \includegraphics[trim={0cm 35mm 4cm 0mm},width=0.7\linewidth,clip]{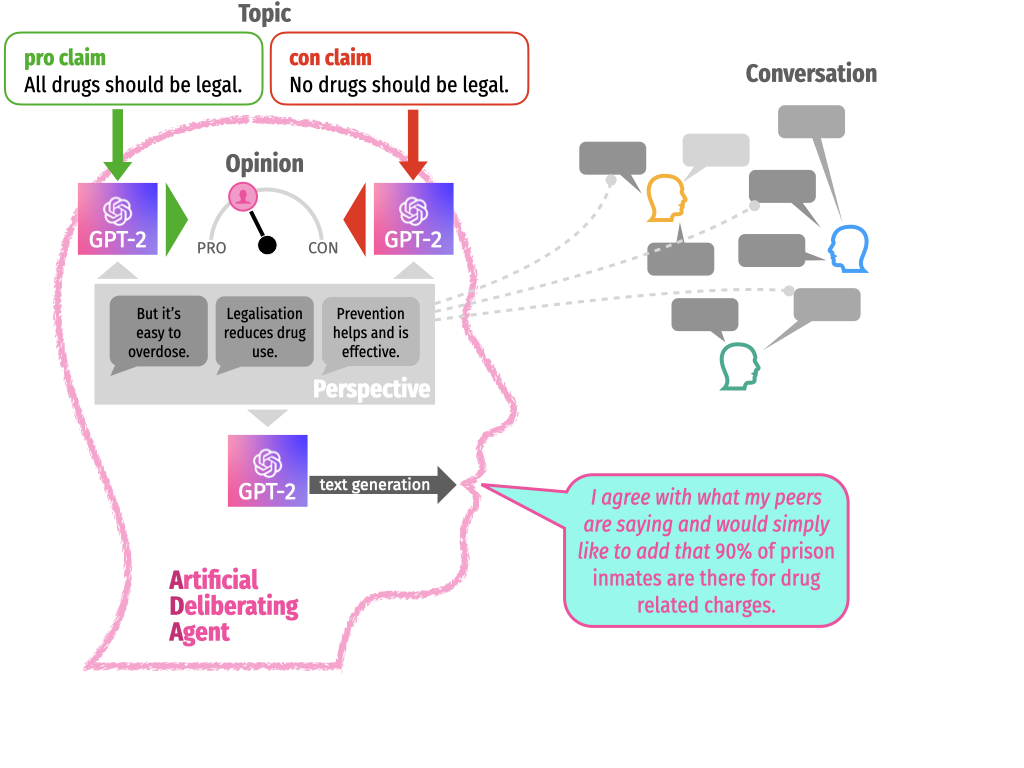}
  \caption{Basic design of artificial deliberative agents ({\smaller ADA}s), which we use to power natural-language agent-based models of argumentation.}
  \label{fig:ADA_illustrated}
\end{figure}

This paper presents a computational model of argumentation that is decidedly not formal in the following sense: It is not built from abstract representations of arguments, but from the very natural language arguments and claims (which are merely \emph{represented} in formal models) themselves. Our natural-language {\smaller ABMA} directly processes and runs on English sentences. A key component of our natural-language {\smaller ABMA} is what we call the artificial deliberative agent ({\smaller ADA}), which we construct with the help of game-changing NLP technology recently developed in AI and computational linguistics \citep{Vaswani2017AttentionIA,Devlin2019BERTPO,Radford2019,brown2020language}. Our design and study of {\smaller ADA}s bears similarities to research on dialogue systems \citep{Zhang2020DialoGPTLG,Bao2020PLATOPD} and chatbots \citep{Adiwardana2020TowardsAH} powered with neural language models; however, unlike these neural dialogue systems, {\smaller ADA}s are equipped with additional cognitive architecture, in particular a minimalist belief system. As illustrated in Figure~\ref{fig:ADA_illustrated}, {\smaller ADA}s have a limited (and changing) perspective of a conversation, which determines their opinion vis-à-vis the central claims of the debate. In addition, {\smaller ADA}s may contribute to a conversation by generating novel posts conditional on their current perspective.  

Now, what is the motivation for developing {\smaller ADA}s and natural-language models of argumentative opinion dynamics in the first place?

A first motive for studying natural-language {\smaller ABMA}s is to de-idealize formal models and to test their results' structural robustness. If, for example, groups with over-confident agents typically bi-polarize in formal models but not in their natural-language counterparts, the original result is not robust and ought to be treated with care. 

A second motive is to "reclaim new territory" by computationally investigating novel phenomena that have not been (and possibly cannot be) represented by formal models. Metaphorical language \citep{hesse1988cognitive}, slurs \citep{rappaport2019communicating}, framing effects \citep{GrueneYanoffRaU2016}, or the invention of entirely new arguments \citep{walton2019computational} is difficult to represent in formal models, but relatively easy in natural-language ones.

A third motive is to create computational models with implicit, natural semantics. Formal models of deliberation cannot escape assumptions about the "semantics" and "logic" of argument, specifically the evaluation of complex argumentation. These assumptions concern, for instance, whether individual reasons accrue by addition, whether the strength of a collection of reasons is merely determined by its weakest link, whether undefended arguments are universally untenable, whether every argument can be represented by a deductive inference, or whether our non-deductive reasoning practice is governed by probabilistic degrees of beliefs. In other words, formal models of argumentative opinion dynamics inevitably rest on highly contested normative theories. With natural-language {\smaller ABMA}s, however, there is no need to take an explicit stance regarding these theoretical issues, because the neural language model, which underlies the {\smaller ADA}, comes with an implicit semantics of argument and takes care of argument evaluation itself. That's why natural-language {\smaller ABMA}s may turn out to be neutral ground and a common point of reference for formal models from rivaling theoretical paradigms. 

A fourth motive is to close the gap between computational simulations on the one side and the vast amount of linguistic data about real conversations on the other side. As natural-language {\smaller ABMA}s do not process mathematical representations of text, but text itself, it is much more straightforward to apply and test these models on text corpora (we'll come back to this in the concluding section).

A fifth and final motive for studying natural-language {\smaller ABMA}s is to explore the capabilities of neural language models. It seems there is currently no clear scientific consensus on what to make of these AI systems. On the one hand, performance metrics for NLP benchmark tasks (translation, text summarization, natural language inference, reading comprehension, etc.) went literally off the chart with the advent of neural language models. On the other hand, some performance gains have been shown to be spurious, as they were just triggered by statistical cues in the data; what's more, the suspicion that large neural language models have simply memorized sufficiently many tasks from the Internet looms large. In this context, the ability of neural language models (mediated via {\smaller ADA}s) to engage in and carry on a self-sustaining, sensible conversation about a topic, while allowing {\smaller ADA}s to reasonably adjust their opinions in its course, may provide further evidence for the cognitive potential, if not capacities of neural language models.    

The paper is organized as follows. Section~\ref{sec:model} presents, step-by-step, the outline of our natural-language {\smaller ABMA}, including key features of {\smaller ADA}s, i.e.: the way an agent's opinion is elicited given her perspective, the way an agent chooses peers in a conversation, the way an agent updates her perspective, and the way an agent generates a new contribution. Note that the guiding design principle in setting up the natural-language {\smaller ABMA} is to rebuild formal reason-balancing models of argumentation \citep{MaesFlache2013, Singer:2019oq} -- which stand in the tradition of Axelrod's model of cultural dissemination \citep{Axelrod1997culture} -- as faithfully as possible and to deviate from these models only where the natural language processing technology requires us to do so.     

As further detailed in Section~\ref{sec:experiments}, we run various simulation experiments with the model to test the effects of (i) different updating strategies and (ii) active contributions to a debate. A closer look at an illustrative simulation run (Subsection~\ref{subsec:illustrative_case}) suggests that our model gives rise to meaningful natural language conversations and that, in particular, {\smaller ADA}s respond to changes in their perspective in a sensible way (as regards both opinion revision and active generation of further posts). Our main findings are reported in Subsection~\ref{subsec:polarization}: First of all, the natural-language {\smaller ABMA} with passive agents qualitatively reproduces the results of formal reason-balancing models regarding the effects of updating strategies on group polarization and divergence. This establishes the robustness of the originally observed effects. Secondly, active generation of novel posts heavily influences the collective opinion dynamics---to the extent that properties of the agents \emph{qua authors} totally dominate the evolution of the conversation. So, the natural-language {\smaller ABMA} identifies a mechanism which is not covered in formal models, but which is potentially of pivotal importance for understanding the dynamics of collective deliberation.   

We close by arguing that there are further fruitful applications of the model, which can be naturally extended to account for phenomena such as multi-dimensional opinion spaces, topic mixing, topic changes, digression, framing effects, social networks, or background beliefs (Section~\ref{sec:discussion}). Although we report results of a preliminary sensitivity analysis in Subsection~\ref{subsec:sensitivity_ana} (suggesting our findings are robust), a systematic exploration of the entire parameter space as well as of alternative initial and boundary conditions appears to be a prime desideratum of future research.

\section{Model}\label{sec:model}

\subsection{Basic Design and Terminology}\label{subsec:basic_design}

A \emph{conversation} evolves around a topic, where a \emph{topic} is defined by a pair of central claims that characterize the opposing poles of the conversation. For example, the claims \{"All drugs should be legal.", "Decriminalize drugs!"\} on the one side and \{"No drugs should be legal.", "Drugs should be illegal."\} on the opposite side may define the topic "legalization of drugs."

A \emph{post} is a small (<70 words) natural language message that can be submitted to a conversation. The conversation's timeline contains all posts that actually have been contributed to the debate ($\textsc{posts}$), including their submission date and author. Let $\textsc{posts}_t$, $\textsc{posts}_{\leq t}$ refer to all posts submitted at, respectively at or before, step $t$.    

Agents participate, actively or passively, in a conversation ($\textsc{agents} = \{a_1,...,a_n\}$). Every agent $a_i$ adopts a specific \emph{perspective} on the conversation, i.e., she selects and retains a limited number of posts which have been contributed to the conversation. Formally, $\textsc{persp}^i_t = \langle p_1, p_2, \ldots , p_k \rangle$ with $p_j \in \textsc{posts}_{\leq t}$ for $j=1\ldots k$.

An agent's perspective fully determines her \emph{opinion} at a given point in time, $\textsc{opin}^i_t \in [0,1]$ (see also Subsection~\ref{subsec:opinion_elicitation}).

Every agent  $a_i$ has a (possibly dynamically evolving) \emph{peer group}, formally: $\textsc{peers}^i_t \subseteq \textsc{agents}$ (see also Subsection~\ref{subsec:peer_selection}). As agents update their perspective (see Subsection~\ref{subsec:perspective_updating}), they exchange their points of view with peers only.

\subsection{The Main Loop}\label{subsec:main_loop}

The following pseudo-code describes the main loop of the simulation of a single conversation.

\begin{algorithm}[H]
\SetAlgoLined
 \For{t in [1...$t_\textrm{max}$]}{
     \For{i in \textsc{agents}}{
      determine the peers of agent i ($\rightarrow\textsc{peers}^i_t$)\;
      update the perspective of agent i ($\rightarrow\textsc{persp}^i_t$)\;
      \If{agent i is contributing at t}{
       generate and submit a new post\;
       }
      elicit the opinion of agent i ($\rightarrow\textsc{opin}^i_t$)\;
     }
 }
 \caption{Main Loop of the Simulation}
\end{algorithm}

\subsection{Opinion Elicitation}\label{subsec:opinion_elicitation}

The opinion of agent $a_i$ at step $t$ is a function of $a_i$'s perspective at step $t$. We define a universal elicitation function $\mathbf{O}$ to determine an agent's opinion: 

\begin{eqnarray*}
\textsc{opin}^i_t = \mathbf{O}(\textsc{persp}^i_t)\\
\mathbf{O}: \mathcal{P}(\mathbb{S}) \rightarrow [0,1]
\end{eqnarray*}

where $\textsc{persp}^i_t$ is a sequence of posts. 

Function $\mathbf{O}$ is implemented with the help of neural language modeling technology (see Appendix). First, we transform the posts in the perspective into a single word sequence (basically by concatenating and templating, as described in the Appendix), which yields a natural language query $Q_\mathrm{elic}(\textsc{persp}^i_t)$. We elicit the opinion of the agent regarding the conversation's central claims by assessing, roughly speaking, the probability that the agent generates a pro-claim rather than a con-claim \emph{given} her perspective. More specifically, we calculate the so-called \emph{conditional perplexity} \citep[cf.][78]{Manning:1999lh} of the pro claim / con claim given the query previously constructed (see also Appendix): $$\mathrm{PPL}_\mathrm{GPT2}(\cdot,Q_\mathrm{elic}(\textsc{persp}^i_t)).$$
Note that perplexity corresponds to inverse probability, so the higher a sequence's perplexity, the lower its overall likelihood, as assessed by the language model. 

Now, let $\mathrm{PPL}^i_t(\mathrm{pro})$ and  $\mathrm{PPL}^i_t(\mathrm{con})$ denote the mean conditional perplexity averaged over all pro claims, resp.\ con claims, conditional on the agents perspective $\textsc{persp}^i_t$. The opinion of agent $a_i$ at step $t$ is then given by

\begin{equation*}
\mathbf{O}(\textsc{persp}^i_t) = \frac{\mathrm{PPL}^i_t(\mathrm{con})}{\mathrm{PPL}^i_t(\mathrm{pro})+\mathrm{PPL}^i_t(\mathrm{con})}
\end{equation*}

Function $\mathbf{O}$ measures the extent to which an agent leans towards the pro rather than the con side in a conversation, as defined by its central claims (recall: low perplexity $\sim$ high probability). It is a polarity measure of an agent's opinion, and we alternatively refer to the opinion thus elicited as an agent's "\emph{polarity}." 

The mean perplexities ($\mathrm{PPL}^i_t(\mathrm{pro})$, $\mathrm{PPL}^i_t(\mathrm{con})$), however, reveal more than an agent's tendency towards the pro side or the con side in a conversation. If, e.g., both $\mathrm{PPL}^i_t(\mathrm{pro})$ and $\mathrm{PPL}^i_t(\mathrm{con})$ are very large, the agent's perspective is off-topic with respect to the central claims. We define an agent's \emph{pertinence} as 

\begin{equation*}
\mathbf{P}(\textsc{persp}^i_t) = 0.5 (\mathrm{PPL}^i_t(\mathrm{pro})+\mathrm{PPL}^i_t(\mathrm{con}))
\end{equation*}

Measure $\mathbf{P}$ allows us to track whether agents change the topic in the course of a conversation.

\subsection{Peer Selection}\label{subsec:peer_selection}

We explore two peer selection procedures: a simple baseline, and a bounded confidence mechanism inspired by the work of \citet{Hegselmann2002}.

\textbf{Universal (baseline).} Every agent is a peer of every other agent at any point in time. Formally, $\textsc{peers}^i_t = \textsc{agents}$ for all $i,t$.

\textbf{Bounded confidence.} An agent $a_j$ is a peer of another agent $a_i$ at some point in time if and only if their absolute difference in opinion is smaller than a given parameter $\epsilon$. Formally, $\textsc{peers}^i_t = \{a_j \in \textsc{agents}: |\textsc{opin}^i_{t-1} - \textsc{opin}^j_{t-1}|<\epsilon \}$

\subsection{Perspective Updating}\label{subsec:perspective_updating}

Agents update their perspectives in two steps (contraction, expansion), while posts that an agent has contributed in the previous time step are always added to her perspective:  

\begin{algorithm}[H]
\SetAlgoLined
\SetKwProg{Fn}{def}{\string:}{}
 \Fn{perspective\_updating(i,t)}{
      retrieve the old perspective, $\textsc{persp\_new}$ = $\textsc{persp}^i_{t-1}$\;
      randomly drop k posts from $\textsc{persp\_new}$ according to how long each post has been included in the perspective\;
      \If{agent i has been contributing at t-1}{
       add the post generated by agent i at t-1 to \textsc{persp\_new}\;  
       k = k-1
       }
      add k further posts to $\textsc{persp\_new}$ according to \emph{perspective\_expansion\_method}\;
      set new perspective, $\textsc{persp}^i_{t} = \textsc{persp\_new}$\;
 }
 \caption{Perspective Updating, Overview}
\end{algorithm}

The contracted perspective, \textsc{persp\_new}, of agent $a_i$ at step $t$ is expanded with posts from the perspectives of all peers, avoiding duplicates, i.e., the agent selects -- according to her specific updating method -- $k$ eligible posts, with $\textsc{posts}_\textrm{el} = \bigcup_{j\in \textsc{peers}^i_{t-1}}\textsc{persp}^j_{t-1} \setminus \textsc{persp\_new} \subseteq \textsc{posts}_{<t}.$ This kind of perspective expansion is governed by one of the following methods:

\textbf{Random (baseline).} Randomly choose and add $k$ eligible posts ($\subseteq\textsc{posts}_\textrm{el}$) to the perspective. 

\textbf{Confirmation bias (lazy).} First, randomly draw $k$ posts from $\textsc{posts}_\textrm{el}$; if all chosen posts confirm the agent's opinion (given the contracted perspective \textsc{persp\_new}), then add the $k$ posts; else, draw another $k$ posts from $\textsc{posts}_\textrm{el}$ and add the $k$ best-confirming ones from the entire sample (of size $2k$) to the perspective.

\textbf{Homophily (ACTB).} choose a peer $a_j \in \textsc{peers}^i_{t}$ in function of the similarity between the agent's and the peer's opinion; randomly choose $k$ posts from the perspective of peer $a_j$, $\textsc{persp}^j_{t-1}$. 

Note that \emph{homophily (ACTB)}, which mimics the ACTB model by \citet{MaesFlache2013}, evaluates the eligible posts \emph{ad hominem}, namely, based on the opinion of the corresponding peer only, while a post's semantic content is ignored. In contrast, \emph{confirmation bias (lazy)}, which implements 'coherence-minded' updating from the model by \citet{Singer:2019oq}, only assesses the eligible posts' argumentative role, irrespective of who actually holds the post. Moreover, we have implemented a "lazy" version of \emph{confirmation bias}, as described above, for computational reasons: a confirmation-wise assessment of \emph{all} eligible posts is practically not feasible.

A full and more precise description of the perspective expansion methods is given in the Appendix.

\subsection{Text Generation}\label{subsec:generation}

Causal language models like GPT-2 are essentially probabilistic next-word prediction machines. Given an input sequence of words $x_1 ... x_k$, the language model predicts---for all words $w_i$ in the vocabulary---the probability that $w_i$ is the next word in the sequence, $\textrm{Pr}(x_{k+1}=w_i|x_1 ... x_k)$. It is obvious that such conditional probabilistic predictions can be used to generate a text word-by-word, and there exist various ways for doing so. This kind of text generation with statistical language models is commonly referred to as \emph{decoding}, and it represents a research field in NLP in its own \citep[c.f.][]{holtzman2019curious, Welleck2020NeuralTG}. Pre-studies have suggested to use randomized beam search (with nucleus sampling) as decoding algorithm (see also Appendix). The key parameters we use to control decoding are 

\begin{itemize}
    \item {\tt temperature}, which rescales the predicted probabilities over the vocabulary (increasing low and decreasing high probabilities if temperature is greater than 1);
    \item {\tt top\_p}, which restricts the set of eligible words by truncating the rank-ordered vocabulary (let the vocabulary be sorted by decreasing probability, and let $r$ be the greatest rank such that the probability that the next word will be $w_1$ or \ldots or $w_r$ is still below {\tt top\_p}, then only $w_1$ \ldots $w_r$ are eligible for being inserted).  
\end{itemize}

In the experiments, we explore the following two decoding profiles:

\begin{center}
    
\begin{tabular}{lcc}
    \toprule
profile & {\tt temperature} & {\tt top\_p} \\
     \midrule
narrow & 1.0 & 0.5 \\
creative & 1.4 & 0.95 \\
    \bottomrule
\end{tabular} 

\end{center}

Metaphorically speaking, the narrow profile emulates a conservative, narrow-minded author who's sticking with the most-obvious, common, usual, and most likely options when writing a text. The creative profile, in contrast, characterizes an author who is much more willing to take surprising turns, to use unlikely phrases and unexpected sentences, who is easily carried away, prone to digress, and much harder to predict.   

Pre-studies show that conversations are extremely noisy if each agent generates and submits a post at every time step; in the following, the probability that an agent is contributing a novel post at a given time step is set to 0.2.

\section{Experiments}\label{sec:experiments}

\subsection{Initialisation}\label{subsec:initialisation}

To run simulations with our natural-language {\smaller ABMA}, the initial perspectives of the agents ($\textsc{persp}^i_0, i=1\ldots n$) have to contain meaningful posts that fall within the conversation's topic. Additionally, it seems desirable that the initial perspectives give rise, group-wise, to a sufficiently broad initial opinion spectrum. 

To meet these requirements, we define topics that correspond to specific online debates on the debating platform \url{kialo.com}, from where we crawl and post-process potential posts. Post-processing involves filtering (maximum length equals 70 words) and conclusion explication. As we crawl posts from a nested pro-con hierarchy, the argumentative relation of each post to the central pro / con claims (root) can be inferred, which allows us to add, to each post, an appropriate pro / con claim as concluding statement. For example, the post "\emph{If drugs being illegal prevented addiction, there would be no drug addicted person. Thus, there is no prevention by just keeping drugs illegal.}" is expanded by "\emph{So, legalization of drugs is a pretty good idea.}" In order to increase diversity of posts, we expand only half of all the posts retrieved by such a conclusion statement.

The experiments described below are run on the topic of the legalization of drugs, initial perspectives are sampled from 660 posts, of which 442 justify or defend the pro claim (legalization).

\subsection{Scenarios}\label{subsec:scenarios}

We organize our experiments along two main dimensions, namely (i) peer \& perspective updating, and (ii) agent type. 

Regarding peer \& perspective \textbf{updating}, we explore four parameter combinations:

\begin{itemize}
    \item \emph{random}: baseline update rules for peers (universal) and perspective (random);
    \item \emph{bounded confidence}: bounded confidence peer selection and random perspective updating;
    \item \emph{confirmation bias}: universal peers and lazy confirmation bias (for perspective updating);
    \item \emph{homophily}: universal peers and homophily (for perspective updating).
\end{itemize}

Regarding \textbf{agent type}, we distinguish passive, and two types of active (i.e., generating) agents:

\begin{itemize}
    \item \emph{listening}: agents are not generating, they only forget, share and adopt posts that have been initially provided;  
    \item \emph{generating narrow}: agents can generate posts, text generation is controlled by the narrow decoding profile;
    \item \emph{generating creative}: agents can generate posts, text generation is controlled by the creative decoding profile. 
\end{itemize}

So, all in all, the simulations are grouped in $4\times 3$ scenarios. For each scenario, we run an \emph{ensemble} of 150 individual simulations. (An online demo will be made available for inspecting the ensemble results.)

\section{Results}\label{sec:results}


\subsection{An Illustrative Case}\label{subsec:illustrative_case}

In this subsection, we present an illustrative simulation run and follow a single {\smaller ADA} during a brief episode of the conversation. The case study is not intended to be representative. Its purpose is two-fold: (i) to illustrate what exactly is going on in the simulations, and (ii) to demonstrate that the model is not just producing non-sense, by showing that we can interpret the {\smaller ADA}'s opinion trajectory as an episode of reasonable belief revision.

\begin{figure}
  \centering
  \begin{tabular}{@{}c@{}c@{}} 
  \includegraphics[height=0.31\linewidth]{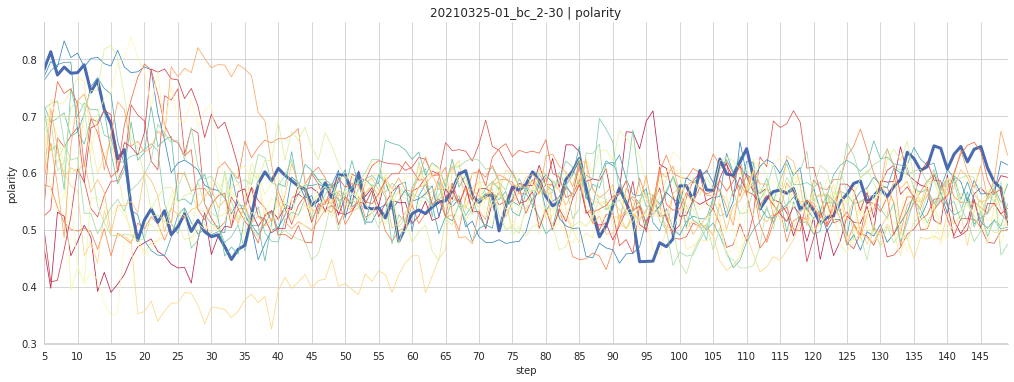} & 
  \includegraphics[height=0.31\linewidth]{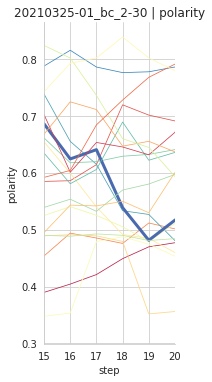} \\
    \small\textit{(a)} & \small\textit{(b)} \\
\end{tabular}
  \caption{Opinion dynamics (polarity) in an illustrative simulation run (20 agents, context size 8, \emph{bounded confidence / generating creative} scenario), (a): full range, (b): zoom into steps 15-20.}
  \label{fig:opdyn_illustrativecase}
\end{figure}

Figure~\ref{fig:opdyn_illustrativecase} plots the opinion trajectories (polarity) of 20 agents over the entire simulation run. Agents select peers in accordance with the bounded confidence mechanism ($\epsilon=0.04$). After the initialisation phase (steps 0--4), the collective opinion spectrum ranges from 0.4 to 0.8. Opinion diversity stays high for the next 20 steps, when the collective profile starts to collapse and more and more agents settle on an opinion around 0.55. From step 65 onwards, a noisy equilibrium state has seemingly been reached.  

Figure~\ref{fig:opdyn_illustrativecase} highlights the opinion trajectory of agent $a_8$. One of its outstanding features is that agent $a_8$ holds an initial perspective that induces a strong opinion pro legalization ($\textsc{opin}^8_5 = 0.8$). In steps 12--20, however, agent $a_8$ completely reverses her opinion. We will now try to make sense of this drastic opinion reversal in terms of changes in the agent's perspective. We limit our discussion to steps 17--19, during which the opinion falls from 0.64 to 0.54 and further to 0.49 (see also Figure~\ref{fig:opdyn_illustrativecase}b).    

In the following display of the agent's perspectives (starting with step 17), posts are highlighted according to whether they \textbf{have been generated by the agent}, \textcolor{teal}{are newly added to the perspective at this step}, \textcolor{teal}{\ul{\textbf{have been generated by the agent at the previous step and are hence newly added}}}, or \textcolor{purple}{will be removed from the perspective at the next step}.

At step~17, the agent's perspective contains a majority of posts plainly pro legalization (17-1, 17-3, 17-5, 17-6, 17-7, 17-8) and no clear reasons against legalization (and that's how it should be, reckoning that the agent comes from the extreme pro side and has -- thanks to bounded confidence -- exchanged posts only with like-minded, pro-legalization peers):

\begin{footnotesize}
\begin{tabular}{@{}p{\linewidth}@{}}
\toprule
 {\scriptsize [17-1]} \textcolor{purple}{Professional addiction treatment is usually [expensive](https://www.addictioncenter.com/rehab-questions/cost-of-drug-and-alcohol-treatment/). I believe all drugs should be legal.} {\scriptsize [17-2]} The term 'increase public health' is subjective. What does that mean? {\scriptsize [17-3]} Marijuana use in the Netherlands has [not increased](https://www.opensocietyfoundations.org/voices/safe-and-effective-drug-policy-look-dutch) following decriminalisation; in fact, cannabis consumption is lower compared to countries with stricter legislation such as the UK. {\scriptsize [17-4]} \textcolor{purple}{It might be a good idea to limit the sale of drugs to adults over the age of 18, and to state clearly that the possession and use of alcohol and cannabis by minors is prohibited.} {\scriptsize [17-5]} Legalising drugs related to date rape could bring the issue further into the public eye, allowing for more widespread education on the topic. {\scriptsize [17-6]} \textbf{The current system is not working. It's absurd to lock people up for using drugs that they choose to make themselves. If they wanted to get high, they'd do it somewhere else.} {\scriptsize [17-7]} If someone wants to go to the supermarket and pick up a few cakes, why shouldn't they? Why shouldn't they be allowed to do so? {\scriptsize [17-8]} \textcolor{teal}{People should be able to ingest anything they want without getting in any trouble for it.}
\\ \bottomrule
\end{tabular}
\end{footnotesize}

The newly added post 17-8 represents a reason pro legalization, which might explain the slight increase of polarity compared to to step~16. Marked for removal in step~18 are: 17-1, an explicit pro reason, and 17-4, a rather nuanced statement which advocates a differentiated policy. Here is how these posts are replaced (cf. 18-7 and 18-8):

\begin{footnotesize}
\begin{tabular}{@{}p{\linewidth}@{}}
\toprule
{\scriptsize [18-1]} The term 'increase public health' is subjective. What does that mean? {\scriptsize [18-2]} Marijuana use in the Netherlands has [not increased](https://www.opensocietyfoundations.org/voices/safe-and-effective-drug-policy-look-dutch) following decriminalisation; in fact, cannabis consumption is lower compared to countries with stricter legislation such as the UK. {\scriptsize [18-3]} \textcolor{purple}{Legalising drugs related to date rape could bring the issue further into the public eye, allowing for more widespread education on the topic.} {\scriptsize [18-4]} \textbf{The current system is not working. It's absurd to lock people up for using drugs that they choose to make themselves. If they wanted to get high, they'd do it somewhere else.} {\scriptsize [18-5]} If someone wants to go to the supermarket and pick up a few cakes, why shouldn't they? Why shouldn't they be allowed to do so? {\scriptsize [18-6]} \textcolor{purple}{People should be able to ingest anything they want without getting in any trouble for it.} {\scriptsize [18-7]} \textcolor{teal}{\textbf{\ul{When you legalize drugs, you're going to have a lot of people who have personal vendettas against certain substances. In this case, the vendettas will probably manifest themselves into violent crime.}}} {\scriptsize [18-8]} \textcolor{teal}{\textbf{According to the Department of Justice, 75\% of the federal prison population is serving time for nonviolent drug crimes. Nearly 90\% of inmates in federal prisons are there for drug crimes.}}
\\ \bottomrule
\end{tabular}
\end{footnotesize}

The post 18-7, which has just been generated by the agent, paints a gloomy (despite somewhat awkward) picture and predicts bad consequences of the legalization of drugs. Post 18-8, which had been previously submitted by agent $a_8$ and then forgotten, is now taken from another peer's perspective and re-adopted by agent $a_8$. It coincidentally picks up the crime trope, claiming that a large proportion of prison inmates have committed drug-related crimes. While 18-8 is, per se, an argumentatively ambivalent statement which can be used to argue both for and against legalization, it's main effect, in this particular context, is apparently to amplify the gloomy outlook cast in preceding 18-7; it hence further strengthens the case against legalization. Given this change in perspective from step 17 to step 18, it makes perfectly sense that the agent's opinion has shifted towards the con side.

Moreover, note that two clear-cut reasons pro legalization are marked for removal (18-3, 18-6), which paves the way for further opinion change towards the con-side.

\begin{footnotesize}
\begin{tabular}{@{}p{\linewidth}@{}}
\toprule
{\scriptsize [19-1]} The term 'increase public health' is subjective. What does that mean? {\scriptsize [19-2]} Marijuana use in the Netherlands has [not increased](https://www.opensocietyfoundations.org/voices/safe-and-effective-drug-policy-look-dutch) following decriminalisation; in fact, cannabis consumption is lower compared to countries with stricter legislation such as the UK. {\scriptsize [19-3]} \textcolor{purple}{The current system is not working. It's absurd to lock people up for using drugs that they choose to make themselves. If they wanted to get high, they'd do it somewhere else.} {\scriptsize [19-4]} If someone wants to go to the supermarket and pick up a few cakes, why shouldn't they? Why shouldn't they be allowed to do so? {\scriptsize [19-5]} \textbf{When you legalize drugs, you're going to have a lot of people who have personal vendettas against certain substances. In this case, the vendettas will probably manifest themselves into violent crime.} {\scriptsize [19-6]} \textbf{According to the Department of Justice, 75\% of the federal prison population is serving time for nonviolent drug crimes. Nearly 90\% of inmates in federal prisons are there for drug crimes.} {\scriptsize [19-7]} \textcolor{teal}{Cocaine is [highly addictive](https://en.wikipedia.org/wiki/Cocaine\_dependence) and easy to become dependent on. I believe legalization of drugs is a really bad idea.} {\scriptsize [19-8]} \textcolor{teal}{It's very easy to overdose on psychoactive substances. It's very difficult to overdose on non-psychoactive substances.}
\\ \bottomrule
\end{tabular}
\end{footnotesize}

The perspective in step~19 newly embraces two posts against legalization, adopted from peers. Post~19-7, in particular, is an explicit con reason, post~19-8 draws attention towards overdosing and hence towards the negative effects of drug use. So, four posts in the perspective now speak against legalization -- compared to 6 pro reasons and no con reason in step 17. Plus, the four con reasons are also the most recent posts (recall that order matters when prompting a language model) and, in a sense, "overwrite" the three previously stated pro claims (19-2 to 19-4). In sum, this explains the sharp opinion change from step 17 to step 19.

\subsection{Global Consensus and Polarization Effects}\label{subsec:polarization}

In this subsection, we characterize and compare the simulated opinion dynamics across our 12 experiments (see Subsection~\ref{subsec:scenarios}), and provide results averaged over the corresponding simulation ensembles.

Based on a cluster analysis (see Appendix for details), we measure the degree of polarization (in terms of clustering coverage and number of clusters), the frequency of bipolarization, and the frequency of full consensus in the simulated conversations. Moreover, we report opinion variance and min-max spread as divergence measures, plus average squared opinion difference -- $(\textsc{opin}_t-\textsc{opin}_{t-1})^2$ -- as volatility measure. Conversations are evaluated at $t=150$.

\begin{figure}
  \centering
  \begin{tabular}{@{}cc@{}} 
  \includegraphics[width=0.4\linewidth]{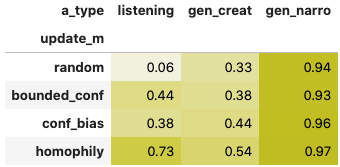} & 
  \includegraphics[width=0.4\linewidth]{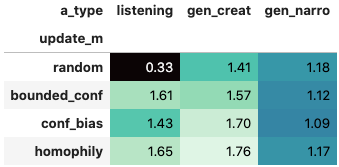}\\
    \small\textit{(a) clustering coverage} & \small\textit{(b) number of clusters} \vspace{4mm} \\ 
  \includegraphics[width=0.4\linewidth]{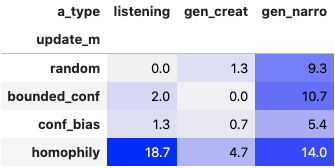} & 
  \includegraphics[width=0.4\linewidth]{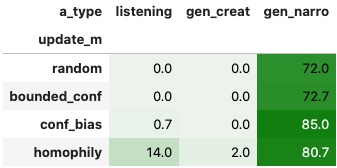}\\
    \small\textit{(c) bipolarization ratio} & \small\textit{(d) full consensus ratio} \\
\end{tabular}
  \caption{Clustering metrics: clustering coverage, number of clusters, frequency of bipolarization (in percent), frequency of full consensus (in percent).}
  \label{fig:results_clustering}
\end{figure}

Figure~\ref{fig:results_clustering} shows how clustering metrics vary across the $4\times 3$ scenarios. Consider, firstly, passive agents who are only sharing but not generating novel posts (column \emph{listening}). With the baseline update mechanism (row \emph{random}), 6\% of the agents fall within an opinion cluster (Figure~\ref{fig:results_clustering}a), and there exists, on average, one cluster in one out of three conversations (Figure~\ref{fig:results_clustering}b). Clustering is much more pronounced for the alternative update mechanisms, with \emph{homophily} in particular attaining more than 70\% coverage and 1.65 clusters per conversation. Let us say that a conversation is in a state of \emph{bipolarization} (\emph{full consensus}) if and only if clustering coverage is greater than 0.9 and there exist two clusters (exists a single cluster). We observe, accordingly, no instances of bipolarization or consensus in the baseline scenario, very few instances for \emph{bounded confidence} and \emph{confirmation bias}, and a significant frequency of bipolarization and full consensus for \emph{homophily} (Figure~\ref{fig:results_clustering}c,d).

This global picture changes entirely as we turn, secondly, to active agents that are generating posts in line with one of the two decoding profiles (cf.\ Subsection~\ref{subsec:generation}). Regarding creative authors (column \emph{gen\_creat}) and irrespective of the particular update mechanism, clustering coverage is between 0.3 and 0.6, there exist approximately 1.5--2 clusters, we observe bipolarization in up to 5\% and full consensus in less than 2\% of all conversations. So we find, compared to passive agents, much stronger clustering in the baseline scenario but significantly less clustering in the \emph{homophily} scenario. Regarding narrow-minded authors (column \emph{gen\_narro}), however, clustering coverage is greater than 0.9, there exists roughly a single cluster per conversation, bipolarization is frequent, and more than 70\% of the debates reach full consensus.

\begin{figure}
  \centering
  \begin{tabular}{@{}cc@{}}
  \includegraphics[width=0.4\linewidth]{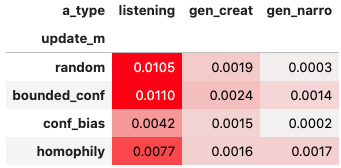} & 
  \includegraphics[width=0.4\linewidth]{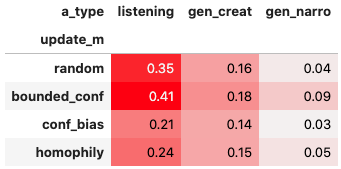} \\
    \small\textit{(a) opinion variance} &     \small\textit{(b) max--min spread}  \\
\end{tabular}
  \caption{Divergence metrics: opinion variance, max-min opinion spread.}
  \label{fig:results_divergence}
\end{figure}

Figure~\ref{fig:results_divergence} describes the extent to which opinions diverge in the 12 simulation experiments. Concerning passive agents, disagreement is most pronounced (both in terms of opinion variance and max--min spread) with \emph{bounded confidence} updating, closely followed by the baseline scenario. Conversations with active agents, in contrast, give rise to much lower levels of disagreement, while narrow-minded authoring is even more agreement-conducive than creative generation, reducing divergence by an entire order of magnitude compared to passive agents.

\begin{figure}
  \centering
  \begin{tabular}{@{}cc@{}} 
  \includegraphics[width=0.4\linewidth]{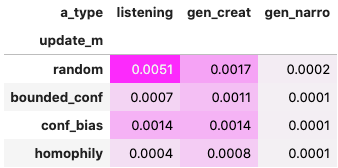} & 
  \includegraphics[width=0.4\linewidth]{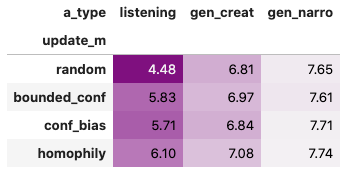}\\
    \small\textit{(a) volatility} &     \small\textit{(b) pertinence}  \\
\end{tabular}
  \caption{Per agent opinion volatility and pertinence.}
  \label{fig:results_volpert}
\end{figure}

As Figure~\ref{fig:results_volpert}a shows, there exist huge differences in terms of per-agent opinion volatility, which is most pronounced for passive agents that randomly adopt novel posts. Structured updating procedures (\emph{bounded confidence}, \emph{confirmation bias}, and \emph{homophily}) have, ceteris paribus, a stabilizing effect and significantly reduce volatility. Creative generation has mixed effects on volatility (depending on the update procedure), while narrow-minded agents possess maximally stable opinions. Finally, Figure~\ref{fig:results_volpert}b reports mean pertinence values for the different scenarios. Recall that pertinence measures the relevance of a perspective for a given pair of central pro/con claims while (cf.\ Subsection~\ref{subsec:opinion_elicitation}): the lower the pertinence value, the more relevant the perspective. Accordingly, agents retain the most relevant perspectives in the baseline scenario. As soon as agents start to generate their own posts, pertinence value increases. However, the conversations stay, on average, faithful to the initial topic (that wouldn't be the case for perplexities above 20, though). Mean pertinence value is, somewhat surprisingly, even slightly lower for creative than for narrow-minded agents.

\subsection{Sensitivity Analysis}\label{subsec:sensitivity_ana}

By varying the number of agents per conversation, the maximum number of posts an agent can hold in her perspective, as well as update-specific parameters (epsilon interval, homophily strength) in the simulation experiments, we obtain a preliminary understanding of the model's sensitivity. Yet, these experiments fall short of a systematic exploration of the entire parameter space, which has not been carried out due to its computational costs, and is certainly a desideratum for future research.

In general, the model seems to yield \emph{qualitatively} similar results when varying key parameters. In particular, structured updating (\emph{bounded confidence}, \emph{confirmation bias}, and \emph{homophily}) with passive agents gives rise to polarization, and once agents get active and start to generate posts, the collective opinion evolution is dominated by decoding parameters -- in the latter case, changes in community or perspective size have quantitatively very little effect.

Regarding \emph{homophily} and \emph{confirmation bias} updating with passive agents, increasing the number of agents per conversation results in more full consensus and less bipolarization. With more agents covering the ground, it seems to be more difficult for subgroups to isolate from other agents and to build a shared, sufficiently distinct perspective. Moreover, increasing the perspective size decreases the frequency of both full consensus and bipolarization, and weakens clustering in general. With more posts in a perspective it takes -- certeris paribus -- longer for an agent to entirely change her perspective; characteristic collective opinion profiles hence build up more slowly and we observe, at a fixed time step, lower polarization and consensus. Plus, it's clear from a look at the time-series that the conversations with \emph{homophily} and \emph{confirmation bias} have not reached a (possibly noisy) equilibrium state at $t=150$, yet. It's therefore a desideratum for future research to run the simulations for much longer time spans.

In particular, epsilon and homophily exponent control \emph{bounded confidence}, respectively \emph{homophily} updating. The model reacts to changes in these parameters as expected: As we increase the epsilon interval, we obtain (with \emph{bounded confidence} updating / passive agents) more clustering and more agreement. Increasing the homophily exponent results (with \emph{homophily} updating / passive agents) in stronger clustering (more consensus, more bipolarization) and greater disagreement.

\section{Discussion and Future Work}\label{sec:discussion}

\paragraph{Structural robustness of formal models.} As regards passive agents, our natural-language {\smaller ABMA} reproduces qualitative results obtained with formal reason-balancing models: \emph{homophily} and \emph{confirmation bias} updating lead to bipolarization, in line with the findings of \citet{MaesFlache2013} resp.\ \citet{Singer:2019oq}. \emph{Bounded confidence} updating increases polarization and disagreement, consistent with \citet{Hegselmann2002}. 
Due to requirements imposed by language modeling technology, the natural-language {\smaller ABMA} is structurally similar to, albeit not identical with the corresponding formal models (e.g., \emph{confirmation bias} implements, for computational reasons, local rather than global search). In addition, the context sensitive, holistic processing of reasons in the natural-language {\smaller ABMA} departs from the strictly monotonic and additive reason aggregation mechanism built into the formal models. All these structural dissimilarities, however, further strengthen the robustness of the findings concerning passive agents.

\paragraph{Limitations of formal models.} 

We have observed that once agents start to generate and submit their own posts, the evolution of the collective opinion profile is dominated by decoding parameters (i.e., properties of the agents as authors). With active agents, we obtain entirely different results for polarization, consensus and divergence than in the experiments with passive agents. In formal reason balancing models, however, agents cannot generate new reasons (or rephrase, summarize, mix, and merge previous ones). So, the natural-language {\smaller ABMA} identifies a potentially pivotal mechanism that's currently ignored by formal models, whose explanatory and predictive scope seems, accordingly, to be limited to conversations and collective deliberations with a fixed set of reasons to share.

\paragraph{Sensitivity analysis.} A systematic sensitivity analysis of the natural language model seems urgent and should go beyond an exploration of the entire parameter space and longer simulation runs. First, some implementation details are worth varying (e.g., the generation of prompts used to query the language model, the post-processing of generated posts, the functional form of the confirmation measure, local search). Second, the initial conditions should be altered, too; in particular, conversations should be simulated on (and be initialized with) different topics.

\paragraph{Empirical applications.} There are multiple routes for empirically applying and testing natural-language {\smaller ABMA}s, which are closed for purely formal models and which might be explored in future work -- as the following suggestions are supposed to illustrate. On the one hand, one can derive and empirically check \emph{macro} predictions of the model: (a) One may test whether groups of conservative authors are more likely to reach full consensus than groups of creative authors. (b) One might try to explain statistical properties of an observed opinion distribution in a debate by initializing the model with posts from that debate and running an entire (perturbed-physics style) ensemble of simulations. (c) Or one might check whether the macro patterns of semantic similarity \citep{reimers-2019-sentence-bert} within a simulated conversation correspond to those in empirical discourse. On the other hand, one can test the \emph{micro} dynamics built into the natural language model: (a) One might verify whether deliberating persons respond to reasons and aggregate reasons in the way the {\smaller ADA} does. (b) Alternatively, one might try to infer -- by means of the natural-language {\smaller ABMA} -- (unobserved, evolving) agent perspectives from (observed) agent contributions so as to account for the agents' (observed) final verdicts on a topic.

\paragraph{Model extensions.} The natural-language {\smaller ABMA} is extremely flexible, as its agents ({\smaller ADA}s) understand and speak English. This allows us to address further linguistic phenomena (slurs, thick concepts) and cognitive phenomena (fallacious reasoning, framing effects) with relative ease, e.g., by systematically changing the prompts used to query the agents, by intervening in a simulated conversation and inserting targeted posts at given step, or by controlling for these phenomena during the initialisation. Likewise, taking opinion pertinence (in addition to opinion polarity) into account in the updating process, eliciting multi-dimensional opinions (with additional pairs of pro-con claims), and mixing multiple topics in one and the same conversation are further straight-forward and easy-to-implement extensions of the model. Obviously, it's also possible to define a neighborhood relation and simulate conversations on social networks. A further set of model extensions regards the heterogeneity of agents: As the model presented in this paper contains (except for their initial condition) identical agents, a first way to increase diversity is to allow for agent-specific (updating and decoding) parameters. Furthermore, we can model background beliefs by fixing immutable posts in an agent's perspective. Finally, there's no reason (besides a computational, practical one) to use one and the same language model to power {\smaller ADA}s; in principle, each agent might be simulated by a specific instance of a language model (with particular properties due to its size, pre-training and fine-tuning) -- plus, these language models might actually be trained and hence evolve in the course of a simulated conversation. The practical effect of this last modification is that agents would display different initial (empty perspective) positions and that an agent might have different opinions at two points in time although she holds one and the same perspective.

\paragraph{Lessons for AI.} This paper has adopted recent technology from \emph{AI and NLP} to advance computational models of argumentative opinion dynamics in the fields of \emph{formal and social epistemology} and \emph{computational social science}. Now, this \emph{might} in turn have repercussions for AI: The fact that we've been able to simulate self-sustainable rational argumentative opinion dynamics suggests that the language model we're using to power agents possesses minimal argumentative capabilities and is, in particular, able to process and respond to reasons in a sensible way. Otherwise, the successful simulation of collective deliberation would be a miracle. Plus, our experiments can be interpreted -- inside out -- as a single agent's attempt to think through a topic by consecutively adopting alternative perspectives (and hence mimicking a deliberation); which suggests that language models are capable of sensible self-talk, consistent with  \citet{shwartz2020unsupervised} and \citet{betz2021thinking}. Finally, such \emph{argumentative} multi-agent systems might be a fruitful design pattern to address tasks in AI and NLP that are difficult to solve with standard systems built around a single agent / a single language model.

\appendix

\section*{Appendix}

\subsection*{Language Model}

In opinion elicitation and text generation, we rely on the pretrained autoregressive language model GPT-2 \citep{Radford2019} as implemented in the \emph{Transformer} Python package by \citet{wolf2019huggingface}.

\subsection*{Prompt Generation}

Let $\textsc{persp}^i_t = \langle p_1 \ldots p_k \rangle$ be the perspective of agent $a_i$ at $t$. To elicit the opinion of agent $a_i$ at $t+1$, we prompt the language model with the following query:

\begin{quote}
    Let's discuss legalization of drugs!
    
    $p_1$
    
    \ldots
    
    $p_k$
    
    I more or less agree with what my peers are saying here. And therefore, all in all,
\end{quote}

When generating a new post at $t+1$, the model is prompted with

\begin{quote}
    Let's discuss legalization of drugs!
    
    $p_1$
    
    \ldots
    
    $p_k$
    
    I more or less agree with what my peers are saying here. Regarding the legalization of drugs, I'd just add the following thought:
\end{quote}

\subsection*{Perplexity}

Let $\mathbf{v}=\langle v_1 \ldots v_k \rangle$ and $\mathbf{w}=\langle w_1 \ldots w_l \rangle$ be two sequences of words. A causal language model (LM) predicts next-word probabilities. Let $p_i$ be the probability that the next word is $w_i$ given the previous sequence $v_1, \ldots, v_k, w_1, \ldots, w_{i-1}$, i.e. 

$$p_i := \mathrm{Prob}_{\mathrm{LM}}(w_i|\mathbf{v}\; w_1\; \ldots\; w_{i-1}).$$

The conditional perplexity of sequence $\mathbf{w}$ given sequence $\mathbf{v}$ is defined as the inverse geometric mean of the predicted conditional probabilities for words in $\mathbf{w}$,

$$\mathrm{PPL}(\mathbf{w}|\mathbf{v}) = \sqrt[k]{\prod \frac{1}{p_i}}.$$

\subsection*{Parameters}

Global parameters of the simulation runs are:

\begin{longtable}{lr}
    \toprule
number of agents & 20 \\
perspective size & 8 \\
maximum steps & 150 \\
relevance deprecation & .9 \\
memory loss (passive) & 1 \\
memory loss (active) & 2 \\
confirmation bias exponent & 50 \\
homophily bias exponent & 50 \\
epsilon & 0.04 \\
    \bottomrule
\end{longtable}

Parameters that control specifically decoding are:

\begin{longtable}{lr}
    \toprule
number of beams & 5 \\
repetition penalty & 1.2 \\
sampling & True \\
    \bottomrule
\end{longtable}

\subsection*{Perspective Updating Methods}

With \textbf{homophily} updating, agent $a_i$ chooses a peer $a_j \in \textsc{peers}^i$ (we drop time indices for convenience) from whom new posts are adopted in function of the similarity in opinion,

$$\mathrm{sim}(i,j) = 1 - |\textsc{opin}^i - \textsc{opin}^j|.$$

The weight agent $a_i$ assigns to peer $a_j$ in randomly choosing her communication partner is further determined by the homophily exponent, hpe:

$$\mathrm{weight}(i,j) = \mathrm{sim}(i,j)^\textrm{hpe}.$$

With \textbf{confirmation bias} updating, agent $a_i$ evaluates eligible posts in terms of their argumentative function. This is modeled by one-sided relevance confirmation, which measures the degree to which a post $p$ confirms the opinion which corresponds to a given perspective \textsc{persp} for an agent $a_i$ at step $t$:

\begin{equation*}
\mathrm{conf}(p) = \begin{cases} 
|\mathbf{O}(\textsc{persp}+p)-\textsc{opin}^i_0| \quad& {\rm if} \quad (\mathbf{O}(\textsc{persp}+p)>\textsc{opin}^i_0)\Leftrightarrow(\textsc{opin}^i_{t-1})>\textsc{opin}^i_0) \quad \\
0 \quad& {\rm otherwise.}
\end{cases}
\end{equation*}

\subsection*{Cluster Analysis}

We carry out the cluster analysis with the help of density based clustering \citep{ester1996density} as implemented in the Python package SciKit learn (setting eps=0.03 and min\_samples=3). As the opinion trajectories are -- depending on the experiment -- very noisy, a clustering algorithm risks to detects merely coincidental clusters that have emerged by chance if it is applied to single data points. In order to identify stable clusters, we therefore apply the clustering algorithm to short opinion trajectories; more specifically, we cluster opinion triples $\langle \textsc{opin}^i_{t-2}, \textsc{opin}^i_{t-1}, \textsc{opin}^i_{t}\rangle$.

\bibliographystyle{plainnat}
\bibliography{bib_all.bib}

\end{document}